\documentclass[conference]{IEEEtran}
\usepackage{subfigure}
\usepackage{listings}
\usepackage[table,xcdraw]{xcolor}
\usepackage{amsmath}
\usepackage{pifont}
\usepackage{graphicx}
\usepackage{amssymb}
\definecolor{mydarkblue}{rgb}{0,0.08,0.65}
\usepackage[colorlinks=true,
    linkcolor=mydarkblue,
    citecolor=mydarkblue,
    filecolor=mydarkblue,
    urlcolor=mydarkblue]{hyperref} 
\usepackage{cleveref}
\usepackage{soul}
\usepackage[shortlabels]{enumitem}
\usepackage{url}
\usepackage{listings}
\usepackage{color}
\usepackage{tikz}
\usepackage{balance}
\usepackage{multirow}
\usepackage{comment}
\usepackage{amsmath}
\usepackage{graphicx}
\usepackage{wrapfig,lipsum,booktabs}
\usepackage{titlesec}
\usepackage[frozencache,cachedir=.]{minted}
\usepackage[symbol]{footmisc}
\usepackage{tablefootnote}
\usepackage{dblfloatfix}
\usepackage{natbib}

\definecolor{codegreen}{rgb}{0,0.6,0}
\definecolor{codegray}{rgb}{0.5,0.5,0.5}
\definecolor{codepurple}{rgb}{0.58,0,0.82}
\definecolor{backcolour}{rgb}{0.95,0.95,0.92}

\makeatletter
\def\blfootnote{\xdef\@thefnmark{}\@footnotetext}
\makeatother

\lstdefinestyle{mystyle}{
  backgroundcolor=\color{backcolour},   commentstyle=\color{codegreen},
  keywordstyle=\color{magenta},
  numberstyle=\tiny\color{codegray},
  stringstyle=\color{codepurple},
  basicstyle=\ttfamily\footnotesize,
  breakatwhitespace=false,         
  breaklines=true,                 
  captionpos=b,                    
  keepspaces=true,                 
  numbers=left,                    
  numbersep=5pt,                  
  showspaces=false,                
  showstringspaces=false,
  showtabs=false,                  
  tabsize=2,
}

\AtBeginDocument{%
  \providecommand\BibTeX{{%
    \normalfont B\kern-0.5em{\scshape i\kern-0.25em b}\kern-0.8em\TeX}}}

\IEEEoverridecommandlockouts
\begin{document}

\makeatletter
  \def\title@font{\Large}
  \let\ltx@maketitle\@maketitle
  \def\@maketitle{\bgroup%
    \let\ltx@title\@title%
    \def\@title{\resizebox{\textwidth}{!}{%
      \mbox{\title@font\ltx@title}%
    }}%
    \ltx@maketitle%
  \egroup}
\makeatother

\title{Zamba: A Compact 7B SSM Hybrid Model}

\author{Paolo Glorioso$ \quad$ Quentin Anthony$ \quad$ Yury Tokpanov$ \quad$  James Whittington$ \quad$  Jonathan Pilault \\ $\quad$  Adam Ibrahim$ \quad$ Beren Millidge \\
{
\small
\{paolo, quentin, yury, james, jonathan, adam, beren\}@zyphra.com
}\\
{}\\
{
 Zyphra
 \small
}\\
{
\small
 Palo Alto, CA
}
}
\maketitle

\setcounter{page}{1}

\begin{abstract}

In this technical report, we present Zamba, a novel 7B SSM-transformer hybrid model which achieves competitive performance against leading open-weight models at a comparable scale. Zamba is trained on 1T tokens from openly available datasets and is the best non-transformer model at this scale. Zamba pioneers a unique architecture combining a Mamba backbone with a single shared attention module, thus obtaining the benefits of attention at minimal parameter cost. Due to its architecture, Zamba is significantly faster at inference than comparable transformer models and requires substantially less memory for generation of long sequences. Zamba is pretrained in two phases: the first phase is based on existing web datasets, while the second one consists of annealing the model over high-quality instruct and synthetic datasets, and is characterized by a rapid learning rate decay. We open-source the weights and all checkpoints for Zamba, through both phase 1 and annealing phases.

\end{abstract}

\section{Introduction}

\begin{figure*}[htbp]
  \begin{center}
  \vspace{-0.5cm}
    \mbox{
      \hspace{-4.0\columnsep}
      \subfigure[Evaluation scores]{
        \includegraphics[width=.45\textwidth,trim=15 -30 2 2,clip]{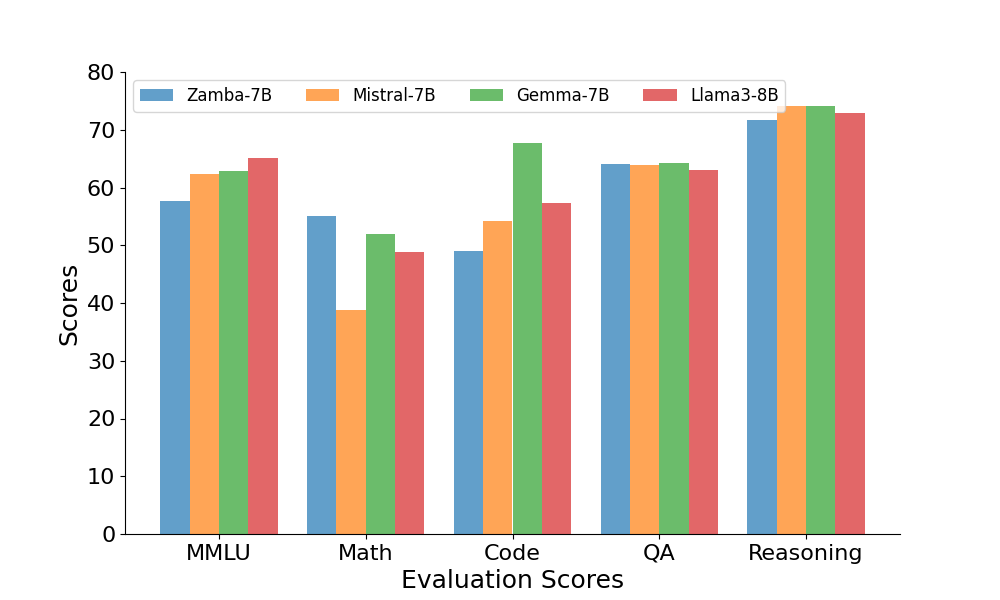}
        \label{fig:evals-bar}
      }
      \hspace{-3ex}
      \subfigure[5-shot MMLU]{
        \includegraphics[width=.35\textwidth,trim=2 2 2 2,clip]{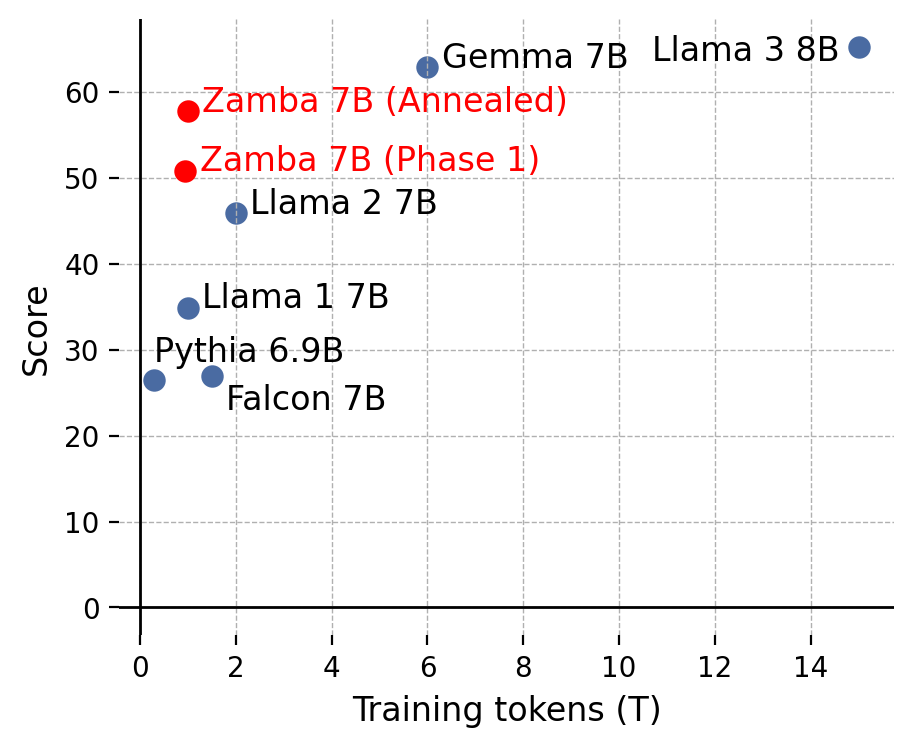}
        \label{fig:mmlu}
      
      }
    }
    \vspace*{-2ex}
    \mbox{
      \hspace{-1.5\columnsep}
      \subfigure[Forward-pass latency at 8k sequence length]{
        \includegraphics[width=.4\textwidth,trim=2 2 2 2,clip]{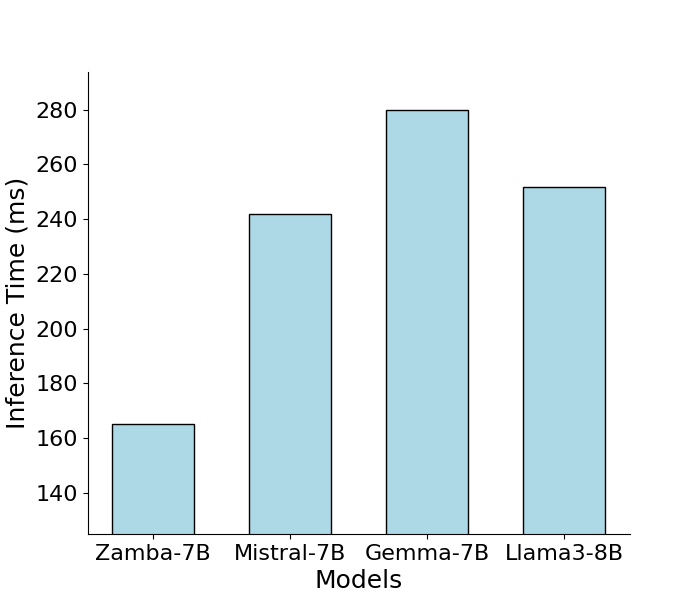}
        \label{fig:inf-lat-bar}
      
      }
      \hspace{3ex}
      \subfigure[Memory usage for generation]{
        \includegraphics[width=.4\textwidth,trim=2 2 2 2,clip]{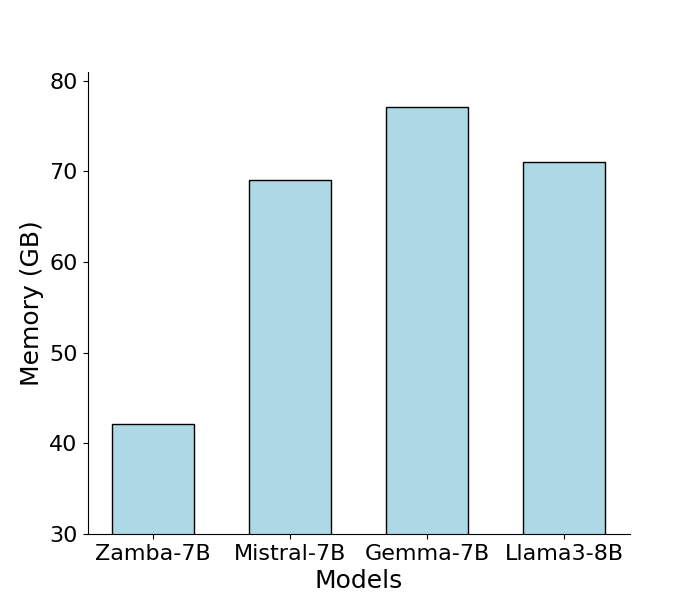}
        \label{fig:gen-mem-bar}
      }
    }
    \vspace*{1\baselineskip}  % Add vertical space between the figures and the caption
    \caption{Evaluation of academic benchmarks, inference and generation performance for Zamba vs other leading models at this scale. We observe that Zamba approaches but is slightly behind the performance of such models at many evaluations, however, as panel (b) shows, it has  been trained on much fewer tokens. Due to its architectural innovations, it can be inferenced significantly faster and requires much less memory for generation. Evaluation results: MMLU is 5-shot, Math is GSM8k 5-shot, code is Human-Eval pass@100, QA is mean of zero-shot BoolQ and OpenBookQA, reasoning is mean of zero-shot ARC, HellaSwag, WinoGrande, and PIQA.}
    \label{fig:fig-1}
  \end{center}
  \vspace*{-1\baselineskip}
\end{figure*}

The transformer architecture \citep{vaswani2017attention} has revolutionized natural language processing and many other fields of deep learning \citep{vit,alphafold2,touvron2023llama} thanks to its ability to scale across model size, dataset size and compute \citep{kaplan2020scaling,gpt3,chinchilla,gopher}.
%The reliable and highly predictable \citep{gpt3,chinchilla,gopher} performance gains arising from scale enable the allocation of significant amounts of effort and capital towards training increasingly large and powerful models for a known and predictable return. 
% However, it is unclear and unlikely that transformers are the only architecture with such favourable scaling properties and it is very likely that superior architectures remain to be discovered.
Nevertheless, it is unclear whether transformers are the only architecture able to leverage scale \citep{bachmann2023scaling}, and new architectures might prove to have even greater scaling potential.
Moreover, the quadratic cost of the core attention computation in transformers remains an important limitation and bottleneck of the architecture. This caveat has led to a search for architectures circumventing this bottleneck while maintaining the performance and scaling capabilities of transformers.

One promising line of research has been in state-space models (SSMs) \citep{gu2021efficiently}, which replaces the attention operation in transformers with a linear dynamical system which can be computed in two ways -- either as a recurrent dynamical system or as a parallel scan. This parallel mode ensures efficient training on GPU hardware, while the recurrent mode which enables linear-time and constant-memory generation, similar to a recurrent neural network (RNN). Because the linear dynamics maintain only a fixed-size hidden state, inference time in these systems is not quadratic in the context length, and does not require a linearly growing KV cache, rendering SSM architectures constant in memory during generation. Potentially, both of these advantages could lead to significantly longer contexts being feasible for SSMs as opposed to transformers.

While early SSM models performed significantly worse than transformers on language tasks, recent SSMs such as Mamba \citep{gu2023mamba} or Griffin \citep{de2024griffin} appear to be closing the gap. Such models rely on input-dependent dynamics for the SSM sequence mixer, analogous to the ways attention uses input-dependent Q, K,V matrices. In spite of these improvements, \citet{jelassi2024repeat} find that they do not fully match the expressivity and performance of transformers at scale, with several works highlighting in particular in-context learning (ICL) weaknesses \citep{park2024mamba,grazzi2024mamba}. Recent works have argued that combining Mamba and attention blocks can lead to improved performance, potentially matching that of full attention (i.e., a pure transformer). In their detailed study of various possible hybrid architectures, \citet{poli2024mechanistic} find that approximately one quarter of the layers being self-attention and the rest SSMs was optimal. %The scalability of Mamba and other SSMs also remains uncertain, with the original Mamba paper training only up to 2.8B parameters. 

Another approach to improving language model efficiency has been Mixture-of-Experts (MoE) architectures \citep{shazeer2016outrageously,fedus2022switch,rajbhandari2022deepspeed}, which can be combined with SSMs \citep{anthony2024blackmamba,lieber2024jamba}. 
% MoE models route to different sets of parameters, usually only of the MLP blocks in transformers, based on the input.
MoE models consist in routing the input to subsets of parameters at specific layers, based on the input to these layers \citep{jacobs1991adaptive}. In transformers, this routing strategy is typically applied to the feed-forward layers (or MLPs) following attention blocks. The intuition being that this routing allows parts of the model to specialize in handling different situations, making it unnecessary to activate all the parameters in the model at all times to achieve good performance. MoEs thus trade increased parameter count---and hence memory cost---for reduced FLOPs during training and at inference. This trade-off is very appealing when serving large language models (LLMs) at scale, contributing to the popularity of MoE architectures. However, in other use cases, e.g. running LLMs on local devices with limited memory such as consumer GPUs, loading the model in memory can become a greater bottleneck than the FLOPs of forward passes. This reverses the trade-off, and it becomes interesting to explore strategies to boost performance using additional FLOPs instead of extra parameters. 

%While LLM performance at the highest level has been and remains dominated by closed-source models requiring prohibitive resources, the smaller model range has become highly democratized with a large number of open-weights models such as Llama \citep{touvron2023llama}, Mistral \citep{jiang2023mistral}, or Gemma \citep{team2024gemma}. This proliferation of open-weight models has resulted in significant advances in our understanding of LLM alignment \textcolor{red}{cite erk, interpretability, and post-training capability enhancements (-/cites)}

Several recent works have begun to openly explore curriculum learning approaches \citep{elman1993learning,bengio2009curriculum} to training large language models. \citet{liu2018curriculum} show that a curriculum of increasing quality of data leads to improvements on question answering tasks. \citet{krishna2023downstream} report that finetuning data appears to be beneficial during pretraining. \cite{gunasekar2023textbooks} demonstrates that continual pretraining with high-quality synthetic data can significantly boost model's performance.
\citet{gupta2023continual} and \citet{ibrahim2024simple} highlight how in continual pretraining settings, strategies with the learning rate schedule and replay lead to improved adaptation on training data, while mitigating forgetting on previously learnt data. The intuition behind Zamba's curriculum approach is that (1) finetuning and synthetic data may be considered to have higher quality than web data, and (2) while high-quality data is much less available than general web data, the fast adaptation in few training steps shown with the last phase of infinite schedules of \citet{ibrahim2024simple} allows leveraging the high-quality data in relatively few iterations at the end of training. This approach has also been hinted at by current frontier models \citep{team2024gemma,team2023gemini} or \citet{abdin2024phi} as a major method of improving performance. During the development of Zamba, many recent works have begun reporting a two-phase curriculum approach, which utilizes standard pretraining web-data for the initial pass followed by an `annealing phase' in which the model is trained on high quality instruct and synthetic data during a rapid learning rate decay. The intuition is that this rapid decay helps the model focus more on assimilating the newer higher quality data than if it was just included throughout all of pretraining, and is especially useful when the amount of high quality data available is small relative to the bulk pretraining data. The first open description of this form of training was miniCPM \citep{hu2024minicpm}, who describe the method in some detail, followed by Nemotron \citep{parmar2024nemotron} and OLMo-v1.7 \citep{olmo-1.7}.

\subsection{Overview}
\label{sec:overview}
\begin{figure}
    \centering
    \includegraphics[width=0.9\columnwidth]{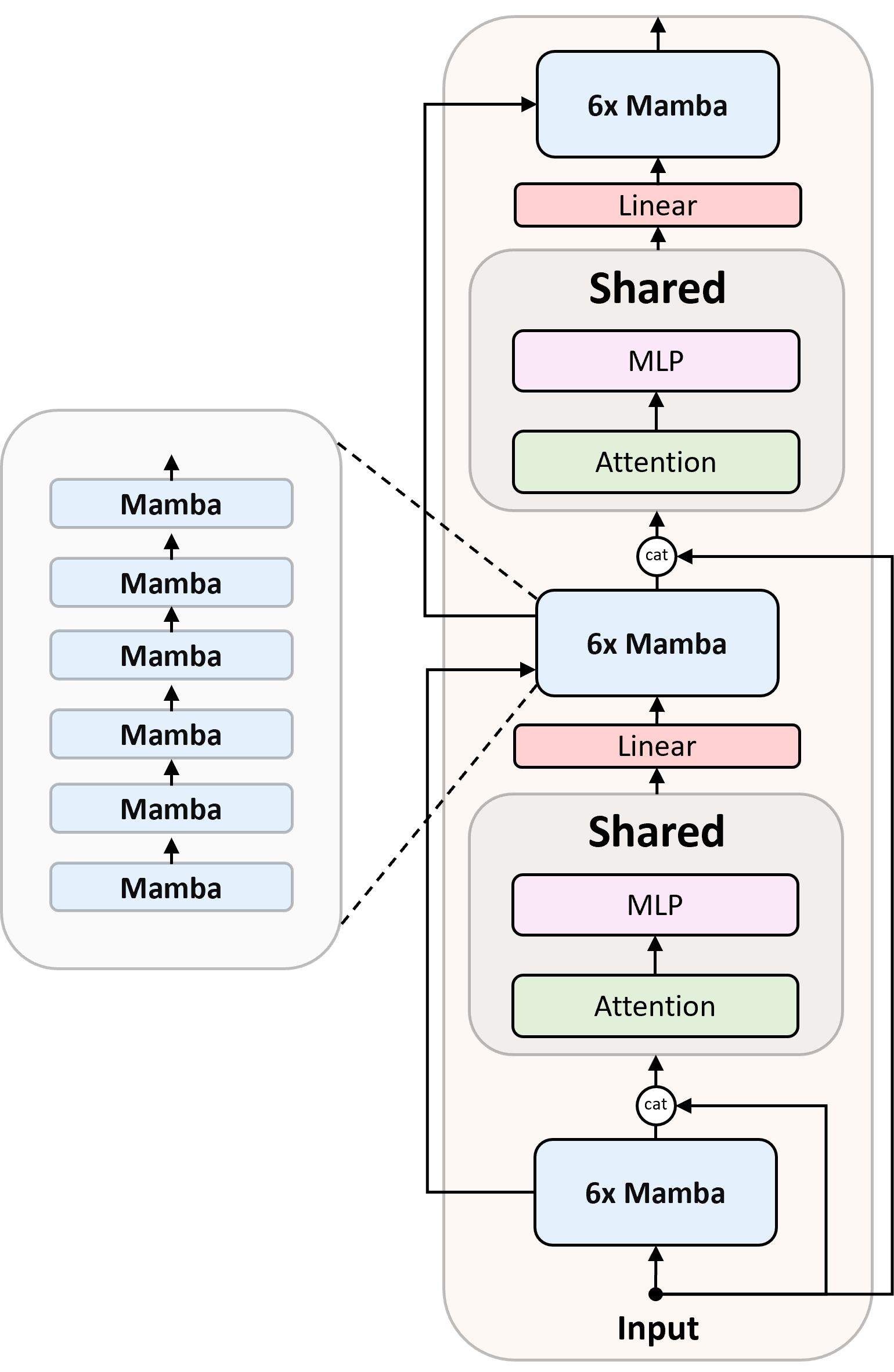}
    \caption{The Zamba architecture. Zamba consists of a backbone of standard Mamba blocks connected to a shared attention and MLP block. This block is repeated every 6 Mamba blocks but has shared parameters, which enables Mamba to utilize more FLOPs for increased performance at the same memory cost. The input embeddings are always concatenated with the residual stream going into the shared attention block as this provides an additional path for the model to remember the inputs. After the block, a learnt linear projection maps the output back to the residual stream.}
    \label{fig:zamba-arch}
\end{figure}

In this technical report, we release and describe the training process for \emph{Zamba}, a 7B Mamba-based SSM with a novel global shared attention architecture. Zamba was trained on only 1T tokens of open web datasets, yet its performance approaches that of the leading $\sim$7B transformer-based models (see Figure~\ref{fig:evals-bar}). Zamba's unique architecture combines a Mamba backbone with a global shared self-attention layer (see Figure~\ref{fig:zamba-arch}), merging the benefits of transformers for retrieval and in-context learning with the inference efficiency of Mamba. This architecture provides a way to boost performance at a small and constant parameter and hence memory cost. Zamba is also the highest-performing SSM in the small 7B model range and the highest-performing dense SSM model available. Zamba matches state-of-the-art 7B models on many linguistic evals, while lagging slightly behind on tests of reasoning and in context learning, which may be due to the significant data disparity between Zamba and other leading $\sim$7B models. 

Our novel architecture was inspired by the relation between the cortex and hippocampus in the brain \citep{whittington_tolman-eichenbaum_2020, whittington_relating_2021} -- where different layers and regions of the cortex, despite performing different tasks, all send and receive information from a single shared memory store in the hippocampus. Additionally, our core concept of shared layers allowing us to spend additional flops to improve performance is closely related to work on recurrent models \citep{dehghani2018universal} which share deep similarities to neural processing \citep{van2020going,tscshantz2023hybrid}. While still early, we believe that the Zamba architecture is an important step towards designing architectures with a global and shared memory store similar to how the brain operates.

We train Zamba in a two-phase approach utilizing a general pretraining and annealing phase on high-quality datasets coupled with rapid learning rate decay. Like \citet{ibrahim2024simple} and in contrast to \citet{hu2024minicpm}, we find it better to rewarm the learning rate followed by a rapid exponential (instead of linear) decay. We find that this significantly improves the performance of our model upon some downstream evaluations, but has little effect on others. To aid open study of the effects of this annealing process, we release both the final annealed model and the model after only phase-1 training. Additionally, we release all checkpoints during training for both phase 1 and the annealing phase. Zamba is the most performant open checkpoints model, surpassing the concurrently released OLMo 1.7~\citep{olmo-1.7} as well as significantly outperforming Pythia~\citep{biderman2023pythia}. We believe that open checkpoints are crucial for allowing the academic community access to key data around state-of-the-art model training and to enable open and productive study of key questions relating to learning dynamics. Our model is also the only open-checkpoints SSM of any significant scale or performance and we hope that our work may be useful for researchers to understand more deeply how learning works in these alternative architectures compared to transformers.

\subsection{Contributions}
\label{sec:contributions}

In sum, our contributions with Zamba are as follows:
\begin{itemize}
    \item State-of-the-art Transformer-SSM hybrid architecture at 7B scale, that preserves the FLOP-efficiency of SSM as well as the in-context learning ability of attention
    \item Novel neuroscience-inspired shared-attention optimization that preserves the modeling performance of independent attention blocks, while saving memory
    \item Successful replication of two-phase training methods in a large-scale model
\end{itemize}

The phase 1 and annealing checkpoints of Zamba can be downloaded from: \href{https://huggingface.co/Zyphra/Zamba-7B-v1-phase1}{https://huggingface.co/Zyphra/Zamba-7B-v1-phase1} and \href{https://huggingface.co/Zyphra/Zamba-7B-v1}{https://huggingface.co/Zyphra/Zamba-7B-v1}.

\section{Model}

Transformer architectures are typically constructed by alternating blocks of self-attention, which perform sequence mixing, and MLPs, which perform per-token processing. These blocks are arranged around a residual stream, which theoretically enables faithful signal propagation through the network, and the outputs from the residual stream are normalized by a layer-norm layer. A schematic mathematical representation of a single transformer layer is as follows:
\begin{align}
    x_{l+1} = x_l +  \text{MLP}(\text{LN}(\text{Self-Attention}(\text{LN}(x_l)))
\end{align}
where $x_l$ represents the activations of the residual stream at layer $l$. The computation and memory requirements and the self-attention block are linear and quadratic in the sequence length, respectively, thus motivating the search for alternative architectures.% such as SSMs which can theoretically handle longer context efficiently.

By contrast, SSM architectures such as Mamba typically collapse the sequence mixing and token processing together into a single block, thus resulting in an entirely uniform architecture. Like a transformer, Mamba also uses a residual stream gated by a layer norm to provide the inputs for each mamba block, resulting in the following schematic:
\begin{align}
    x_{l+1} = x_l + \text{Mamba}(\text{LN}(x_l))
\end{align}

Like all SSMs, the sequence mixer of the Mamba block is based around a core linear dynamical system. The key innovation of Mamba is to make the control and observation matrices, as well as the timestep of this system, input-dependent. That is, the core dynamics of the Mamba SSM are, schematically,
\begin{align}
    h_{t+1} &= \text{exp}(A\delta_t) h_t  + B_tx_t  \nonumber \\
    y_t &= C_t h_{t+1} \label{mamba_seq_mix_eqn}
\end{align}
where $x_t$ is the input to the system at each sequence element, $h_t$ is the internal `memory' state of the Mamba block, $y_t$ is the output of the sequence mixer, $\delta_t$ is an input-dependent timestep arising from the discretization of the continuous time dynamics, and $B_t$ and $C_t$ are input-dependent matrices, while $A$ is an input-independent matrix.

The Mamba block performs both sequence mixing through the core SSM components as well as token processing via projection layers and a gating unit.
Inputs to the Mamba block are first processed by a linear layer which splits apart the input into a vector to be processed and a gating vector. This is followed by a 1d convolution which performs some preliminary sequence mixing, the output of which is then used to create the input-dependent $B_t$,$C_t$ and $\delta_t$ matrices, followed by the SSM sequence mixer as in Equation \ref{mamba_seq_mix_eqn}. The output $y_t$ of the sequence mixer is then elementwise multiplied by the gating vector from the input, before being passed through a final linear layer to the output. 
Schematically,
\begin{align}
    \text{Mamba}(x) = \text{Lin} (\sigma(\text{Lin}(x)) \odot \text{SSM}(\text{Conv1D}(\text{Lin}(x))))
\end{align}
Due to the gating and the linear projections, the Mamba layer can perform token processing as is performed by the MLPs in the transformer architecture, while also containing the SSM sequence mixer. This means that in a Mamba vs Transformer architecture of approximately the same size, the Mamba model will have twice the sequence mixers, although potentially they are less expressive than full self-attention.

Zamba consists of a pure Mamba backbone augmented with an additional global shared self-attention (GSA) block every $N$ Mamba blocks. The GSA block consists of a self-attention block and MLP block in series, with shared weights for both. This means that although attention is performed many times throughout the network, it uses the same parameters and thus reduces the memory cost of the model both for the attention parameters and, crucially, for the KV cache size during generation. To allow for some specificity for each GSA call, there is an un-shared learnable linear mapping from the GSA block to the residual stream.
Additionally, each input to the GSA block is the concatenation between the residual activations at that layer with the initial residual activities (i.e., after the initial input embedding). This was done to ensure that information from the start of the network is always available to the self-attention. We find that this gives a consistent improvement in performance, perhaps indicating that the residual stream does not maintain information as faithfully throughout the network as the network requires. Due to concatenation, query, key and value vectors in the GSA block have double dimension, which is then projected back, prior to the MLP block, to the residual dimension through the attention's out projector. %We concatenate the residual activations with the original activations as input to the GSA block. 
%Since the output of the GSA block is thus of dimension twice the residual dimension, we add an additional learnable linear layer to project back to the residual dimension. 
Schematically, the Zamba GSA block can be represented as,
\begin{align}
    y_{l} = \text{Lin}_{l} ( \text{MLP}(\text{LN}(\text{Self-Attention}(\text{LN}([x_l, x_0])))))
\end{align}
every $N$ blocks and otherwise pure Mamba. The subsequent Mamba layer is then computed by adding $y_l$ to the non-residual part of the input:
\begin{align}
    x_{l+1} = x_l + \text{Mamba}(\text{LN}(x_l + y_l))
\end{align}
In early ablation studies on small models, we found that this architecture gave a consistent improvement over pure Mamba and also significantly beat parameter-matched Mambaformer \citep{park2024mamba}, or hybrid architectures, which combine self-attention and Mamba blocks. As such, the success of the Zamba architecture provides evidence that perhaps many of the parameters of self-attention layers are redundant and that they can be replaced with a single, repeated, self-attention operation. We believe that, given that often the primary bottlenecks for running models is the memory cost of the model both in number of parameters and generation, it is valuable to investigate architectures which trade additional inference FLOPs for reduced parameters while maintaining performance. 

\subsection{Phase 1}

We trained Zamba-v1 in two phases. First, we performed a standard pretraining phase on open web datasets. During this phase, Zamba-v1 was trained on approximately 950 Billion tokens. We performed a cosine decay from an initial learning rate of $1.5*10^{-4}$ to $7.5*10^{-5}$; this decay is significantly less steep than is common, to ensure that there is room for the rapid decay in the annealing phase. Zamba-v1 was trained on 128 H100 GPUs for approximately 30 days. We used data-parallelism and tensor-parallelism across two ranks, performed across attention, MLP and the Mamba blocks. We used the Zero-1 distributed optimizer to shard optimizer states across ranks and activation checkpointing. We trained with a sequence length of 4096 tokens, using the Mistral tokenizer. A full set of training hyperparameters can be found in Appendix \ref{model_training_hyperparams_section}.

\subsection{Datasets}

Our phase 1 dataset consists of a mixture of open web datasets and included The Pile \citep{gao2020pile}, RefinedWeb \citep{penedo2023refinedweb}, C4 \citep{raffel2020exploring}, PeS2o \citep{peS2o}, and arxiv. The composition of our dataset is shown in Table \ref{table:dataset_proportions}.

\begin{table}[h]
\centering
\begin{tabular}{|>{\columncolor[HTML]{EFEFEF}}c|c|}
\hline
\cellcolor[HTML]{C0C0C0}\textbf{Dataset Name} & \cellcolor[HTML]{C0C0C0}\textbf{Proportion} \\
\hline
peS2o & 4.89\% \\ %0.04892824535
\hline
C4 &  12.85\% \\ %0.1284899808
\hline
Pile & 15.48\% \\ %0.154583177
\hline
RefinedWeb & 62.35\% \\ %0.6235465495
\hline
Cosmopedia & 3.55\% \\ %0.03548190327
\hline
arxiv & 0.90\% \\ %0.008970143972
\hline
\end{tabular}
\vspace{0.5cm}
\caption{Proportion of tokens of sub-datasets making up our full 1T pretraining dataset. All datasets we use are open and available.}
\label{table:dataset_proportions}
\end{table}

We performed relatively minor filtering and then LSH-based fuzzy deduplication on this dataset. 

We applied the following filters to Pile, and C4-en components of our datasets: minimum length of 100 characters, minimum mean word length of 3 characters, maximum mean word length of 12 characters, maximum fraction of non-alphanumeric characters of 0.3, maximum fraction of numeric characters of 0.2. For fuzzy deduplication we computed minhashes with signature size of 128, computed on 13-grams based on words. This filtering removed roughly 10\% of Pile-dedup and 0.5\% of C4-en. 

We deduplicated each dataset both against itself and against the other datasets in our full dataset. We built the LSH index targeting 50\% Jaccard similarity threshold (25 bands with the range of 5) by inserting the following datasets in the particular order: first Pile-dedup, then C4-en, peS2o, arxiv\_s2orc\_parsed. 
We did not insert RefinedWeb into the index, but simply ran its every document through it (RefinedWeb is extensively deduplicated by their authors, so we decided not to spend time to deduplicate is against itself, but only against other datasets). The percentage of identified duplicates in every dataset can be found in Table \ref{table:dataset_dedup}. We did not apply any processing to Cosmopedia, since we reasoned its synthetic nature should imply it is mostly high quality and not duplicated in other datasets.

We did not upsample or perform multiple epochs over any components of our phase 1 dataset. Our total dataset consisted of just over 1 trillion tokens, of which the pretraining phase utilized 950 billion tokens.

\begin{table}[h]
\centering
\begin{tabular}{|>{\columncolor[HTML]{EFEFEF}}c|c|}
\hline
\cellcolor[HTML]{C0C0C0}\textbf{Dataset Name} & \cellcolor[HTML]{C0C0C0}\textbf{Percentage of duplicates} \\
\hline
Pile & 25\% \\
\hline
C4 & 30\% \\
\hline
peS2o & 31\% \\
\hline
arxiv & 7.1\% \\
\hline
RefinedWeb & 1.2\% \\
\hline
\end{tabular}
\vspace{0.5cm}
\caption{Percentage of documents marked as duplicates by fuzzy LSH-minhash deduplication.}
\label{table:dataset_dedup}
\end{table}

\subsection{Annealing}
The annealing phase follows the intuition from the infinite learning rate schedules presented in \citet{ibrahim2024simple}, which shows quick improvements thanks to an exponential decay. Doing this on higher-quality data allows curriculum approaches to require much less high-quality tokens than usual during pretraining. Similarly to miniCPM \citep{hu2024minicpm} and Nemotron \citep{parmar2024nemotron}, we performed a rapid LR decay over a dataset consisting of a mixture of high-quality tokens and our pretraining dataset. We used a replay fraction of 60\% original pretraining data and 40\% new annealing datasets. Our annealing dataset comprised a large collection (over 100) of existing high-quality datasets from various sources. These include large math (such as StackMathQA) and code datasets (such as EvolInstructCode), as well as instruct finetuning (such as OpenOrca) datasets, and some synthetic data examples from more powerful language models. For most of these datasets, we performed only one epoch, but for select high-quality subsets, we performed 2 epochs. We utilized an exponential decay schedule of the form $\eta_t = A e^{-t/(\gamma T)}+B$, where $t$ is the iteration number, $T$ is the total annealing iterations, and we found $\gamma=0.25$ to be the optimal value. The coefficients $A$ and $B$ are found by requiring $\eta_0$ and $\eta_T$ to be the learning rate after warmup and at the end of annealing, respectively, where we used $\eta_0=1.1*10^{-4}$ and $\eta_T=10^{-7}$. Unlike miniCPM, we found that rewarming the learning rate from $0$ back to the maximum learning rate and then decaying again performed better than beginning the annealing phase at the final learning rate of the original pretraining run -- a similar finding to \citet{ibrahim2024simple} in the continual pretraining setting. We additionally found that our exponential decay schedule outperformed the linear decay used in miniCPM. Finally, while we did see a decrease in loss on the original pretraining dataset, it was not as pronounced as described by miniCPM, which we ascribe to Zamba being a much larger model trained on more tokens, or could potentially highlight differences in our annealing datasets or schedule.

\renewcommand*{\thefootnote}{\fnsymbol{footnote}}

\begin{table*}
\begin{tabular}{|>{\columncolor[HTML]{EFEFEF}}c|c|c|c|c|c|c|c|c|c|}
\hline
\cellcolor[HTML]{C0C0C0}\textbf{Model} & \cellcolor[HTML]{C0C0C0}\textbf{PIQA} & \cellcolor[HTML]{C0C0C0}\textbf{ARC-Easy} & \cellcolor[HTML]{C0C0C0}\textbf{ARC-Challenge} & \cellcolor[HTML]{C0C0C0}\textbf{BoolQ} & \cellcolor[HTML]{C0C0C0}\textbf{WinoGrande} & \cellcolor[HTML]{C0C0C0}\textbf{HellaSwag} & \cellcolor[HTML]{C0C0C0}\textbf{OpenBookQA} & \cellcolor[HTML]{C0C0C0}\textbf{MMLU} & \cellcolor[HTML]{C0C0C0}\textbf{Tokens} \\ \hline
Llama 3 8B & 80.96 & 77.61 & 53.58 & 81.16 & 73.24 & 79.13 & \underline{45.0} & \textbf{65.17} & 15 T  \\ \hline
Mistral 7B v0.1 & \textbf{82.26} & \underline{79.59} & \underline{53.92} & \textbf{83.64} & \underline{73.88} & \textbf{81.07} & 44.2 & 62.30 & ? \\ \hline
Gemma 7B & 81.12 & \textbf{80.77} & \textbf{54.27} & 83.12 & 73.8 & \underline{80.46} & \textbf{45.2} & \underline{62.85} & 6 T \\ \hline
Pythia 6.9B & 76.93 & 63.4 & 35.9 & 64.2 & 62.43 & 65.8 & 38.8 & 26.51 & 0.3 T \\ \hline
Falcon 7B & 80.74 & 70.88 & 43.69 & 73.79 & 67.25 & 76.39 & 43.6 & 26.90 & 1.5 T \\ \hline
Llama 2 7B & 79.05 & 74.58 & 46.16 & 77.68 & 69.06 & 75.97 & 44.2 & 45.88 & 2 T \\ \hline
Llama 1 7B & 79.16 & 72.94 & 44.71 & 75.11 & 70.01 & 76.24 & 44.6 & 34.88 & 1 T \\ \hline
OLMo 7B 1.0 & 79.43 & 68.69 & 40.36 & 72.48 & 66.37 & 75.67 & 42.2 & N/A\footnote[1] & 2.5 T \\ \hline
\textbf{Zamba 7B (Phase 1)} & 79 & 70.74 & 43.9 & 77.5 & 70.8 & 77.14 & 44.2 & 50.82 & 0.95 T \\ \hline
\textbf{Zamba 7B (Annealed)} & \underline{81.37} & 74.5 & 46.48 & \underline{83.6} & \textbf{76.4} & 80.24 & 44.6 & 57.72 & 1 T \\ \hline
\end{tabular}
\caption{Zero-shot evaluation results on language modelling benchmarks, except for MMLU which is five-shot, comparing Zamba with open-weight models of a comparable parameter count. We also report the training token count (if known) to demonstrate the sample efficiency of Zamba. First place is \textbf{bolded}, second place is \underline{underlined}.}
\label{tab:my_label}
\end{table*}

\section{Performance}

\begin{figure}
    \centering
    \includegraphics[width=0.99\linewidth]
    {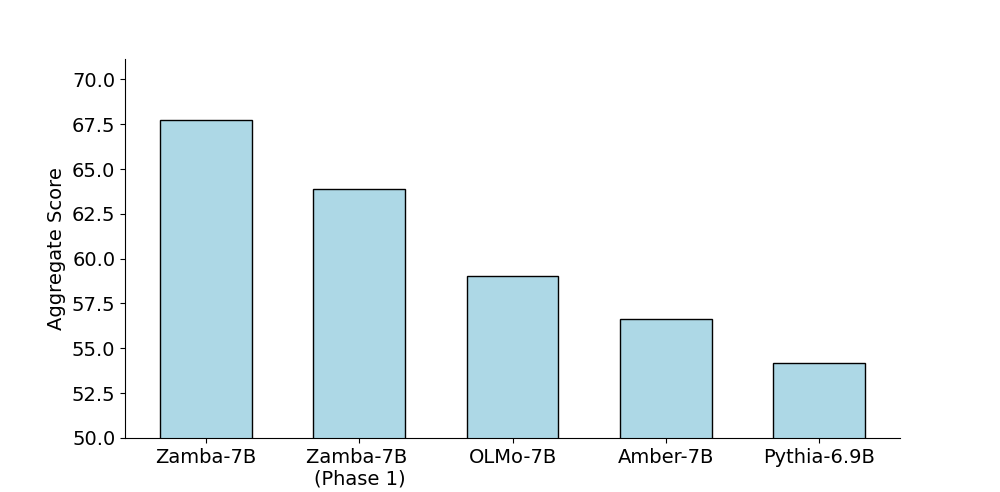}
    \vspace{-0.2cm}
    \caption{Aggregated evaluation performance of Zamba vs alternative open-checkpoint models Olmo 1.0, Amber-7B and Pythia. We observe that Zamba both phase 1 and annealed models significantly outperform competing open-checkpoint models. The aggregate eval is the mean score of PIQA, ARC-Easy, ARC-Challenge, HellaSwag, WinoGrande, BoolQ, and OpenBookQA.}
    \label{fig:open-ckpts}
\end{figure}

We evaluate Zamba on a suite of standard language modelling evaluation benchmarks including reasoning, question-answering, math, code, and general knowledge questions. In general, the trend is that Zamba performs well for its size, and is generally far ahead of competing models trained on open datasets (see Figure~\ref{fig:open-ckpts}), but slightly lags behind the performance of the leading models (see Figure~\ref{fig:fig-1}). However, per-token (and hence per training FLOP), Zamba is extremely efficient relative to comparable models. We believe that this is primarily due to the fact that Zamba has been trained on significantly fewer tokens than leading models, and potentially using lower quality open-web data compared to the secret, in-house datasets of the large AI companies. Nevertheless, we note that Zamba outperforms Llama2 and hence is not a weak model. Zamba's performance on code is relatively poor compared to leading models and we believe that this is most likely explained by the relative lack of code in our dataset, since we used only direct web datasets and not github. Zamba appears to perform well, often competitively with the leading models on general language modeling and reasoning benchmarks such as PIQA, Winogrande, and HellaSwag, which is interesting since it has likely been trained on substantially fewer and potentially lower-quality tokens than these models, and this may be indicative of the architectural benefits of Zamba.

\begin{figure*}[htbp]
  \begin{center}
      \mbox {
          \hspace{-2\columnsep}
          \subfigure[Inference latency.]
          {
              \includegraphics[width=.33\textwidth,trim=2 2 2 2,clip]{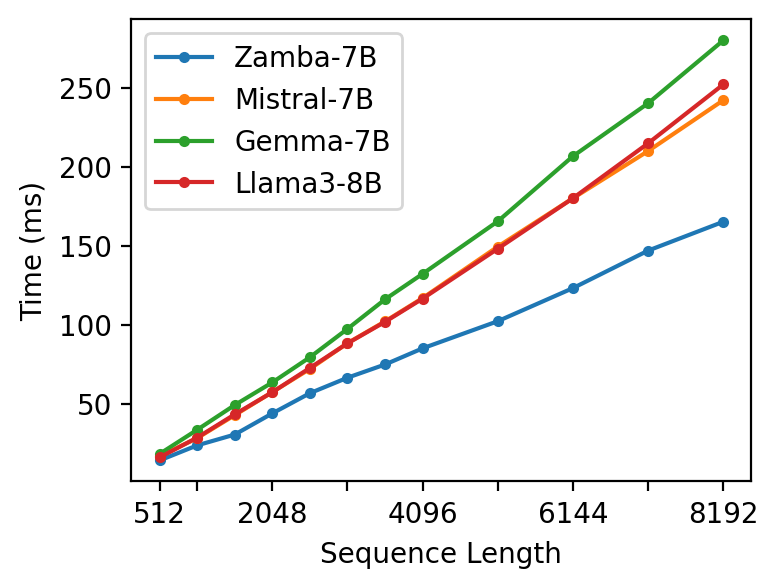}
              \label{fig:inf-lat-line}
          }
          \hspace{-2ex}
          \subfigure[Generation latency. Input sequence length of 2048.]
          {
              \includegraphics[width=.33\textwidth,trim=2 2 2 2,clip]{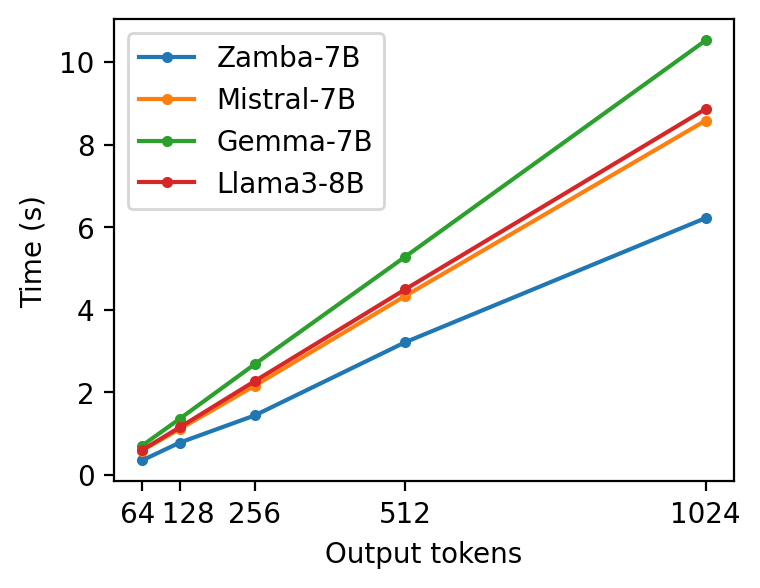}
              \label{fig:gen-lat-line}
          }
          \hspace{-1ex}
          \subfigure[Generation per-GPU memory. 8192 input tokens.]
          {
              \includegraphics[width=.33\textwidth,trim=2 2 2 2,clip]{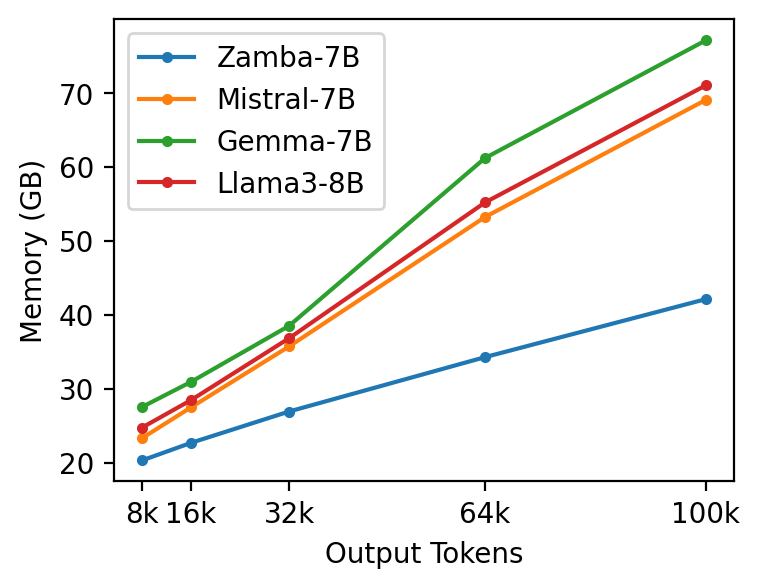}
              \label{fig:gen-mem-line}
          }
      }
      \vspace*{-0.5\baselineskip}
      \caption{Inference/generation results on 2 H100 GPUs using tensor parallelism. All numbers taken with a batch size of 1. Competing models were evaluated using vLLM~\citep{kwon2023efficient}.}
      \label{fig:infer-results}
  \end{center}
\vspace*{-1\baselineskip}
\end{figure*}

In terms of inference and generation efficiency, Zamba is extremely performant. Despite utilizing more FLOPs per parameter due to our parameter sharing, Zamba's forward pass is significantly faster than competing models at the 7B scale, an advantage which increases with longer sequence lengths. Additionally, due to Zamba's SSM backbone, memory required for KV caching in Mamba is reduced by a large factor compared to other models of a similar scale, thus enabling Zamba to generate more efficiently and achieve significantly longer contexts on a single device (see Figure~\ref{fig:infer-results}.

Zamba possesses a number of architectural qualities that make it particularly amenable to inference efficiency:

\begin{itemize}
    \item Zamba's sparing use of attention layers keeps its KV-cache memory particularly low. Specifically, there is a single attention block applied 13 times within Zamba-7B, with independent activations and KV-cache entries at each invocation.
    \item Mamba blocks have a significantly higher throughput than comparable attention or MLP blocks~\citep{gu2023mamba}.
    \item We apply the most efficient available kernel for each available block. Mamba kernels were taken from the open-source mamba implementation~\citep{gu2023mamba} and tuned for the H100 architecture. Attention is implemented via the Flash Attention v2 kernels~\citep{dao2023flashattention2}. The RMSNorms are implemented from the Transformer Engine library~\citep{transformer_engine}. 
\end{itemize}

At the time of writing, Zamba is the best-performing open-checkpoints model available. We believe that Zamba is of significant interest to those studying training dynamics, both because of its performance and because of its unique architecture. By comparing representational dynamics in models such as Zamba with standard transformer architectures such as OLMo~\citep{groeneveld2024olmo}\footnote[1]{We were unable to collect evaluation results for OLMo 1.0 MMLU, nor any results for OLMo 1.7~\citep{olmo-1.7}, due to incompatibilities with the language model evaluation harness~\citep{eval-harness} v0.4.0.}, we hope that progress will be made towards understanding how model architectures affect the formation of representations during training.

\section{Related Work}

\subsection{State-space-models}
The quadratic cost of attention in the sequence length has brought about a number of approaches to try to perform language modeling capabilities of transformers but without this limitation. If this is possible it would enable significantly larger context lengths and much more efficient generation in terms of both FLOPs and memory required to store the KV cache. One line of work aiming to tackle this problem has been utilizing state-space models \citep{gu2021efficiently,gu2020hippo,gu2021combining}. State space models utilize a linear dynamical system approach to model language. Such a dynamical system maintains a constant size `memory' which is maintained throughout the sequence which must encode all the information needed to predict the next token. Unlike RNNs however due to the linearity of the dynamics, SSMs can be written and computed as a convolution or a parallel scan enabling efficient forwarding in a long sequence like a transformer which enables training at scale. The key limitation of SSMs is the restricted form that the sequence mixing can take which enables efficient computation, as well as the compression of all information into a fixed size memory. This renders the expressive power of SSMs less than transformers in theory although it is unclear to what extent natural language requires the full expressive powers that transformers give. While early SSM models performed significantly worse than transformers recent models such as Mamba, and S6 have claimed to perform and scale on par, as have other SSM adjacent architecture such as Griffin~\citep{de2024griffin}, RWKV~\citep{peng2023rwkv,peng2024eagle}, and RetNets~\citep{sun2023retentive}. These models are typically more expressive because they possess selective (input-dependent) gating and control of their internal dynamics, similar to the input dependence of the query, key and value vectors in a transformer. 

Recent work however has highlighted potential deficiencies of pure SSMs at in context learning and other algorithmic tasks and several groups have found that hybridizing SSMs with attention gives performance on par or exceeding transformers cite. While most of this work adds many attention layers, with Zamba we pose and show evidence for an interesting hypothesis — that one attention layer is all you need.

\subsection{Annealing}
While earlier works may have utilized a two-phase annealing schedule, but not disclosed it, the first work that openly discussed in detail a two-phase schedule was miniCPM~\citep{hu2024minicpm}. They describe performing a first slow learning rate decay over the majority of pretraining data, followed by a more rapid decay over high-quality datasets, including instruct data. Since then, several other recently released models have explicitly or elliptically discussed annealing phases. These include Nemotron~\citep{parmar2024nemotron4} and JetMoE~\citep 
{shen2024jetmoe}, both claiming that annealing significantly improved model quality. After Zamba's initial release, OLMo 1.7~\citep{olmo-1.7} also claimed to have performed annealing, which significantly improves the results on MMLU~\citep{mmlu}, and utilized a linear scheduler based on a number of higher-quality dataset, such as wikipedia and flan-instruct. Zamba follows a similar approach, but utilizes a different dataset mix, as well as a faster exponential decay schedule and a learning-rate rewarmup from zero.

\subsection{Open checkpoints}
While releasing models with full checkpoints during training is rare, there are a number of works that have done so. The Pythia suite~\citep{biderman2023pythia} pioneered this approach and has been instrumental in many works on interpretability and training dynamics. However, trained for scientific purposes, Pythia models attain far from state of the art performance. Several more recent works also aim to open and democratize LLM pretraining by offering checkpoints and training details. These include OLMo~\citep{groeneveld2024olmo}, who provide detailed training code and checkpoints for a 7B parameter model, and LLM360~\citep{liu2023llm360}, who provide descriptions of training and open-checkpoints models based on the Llama architecture. Additionally, for study of pure Mamba architectures, Zyphra released open checkpoints of a small 370m pure Mamba model trained on the Pile dataset~\citep{zyphra-mamba}.

\section{Discussion}

In this technical report, we introduce Zamba, a 7B open-source SSM highly competitive with leading models. We thus demonstrate conclusively the scalability of SSM architectures to this scale. Moreover, we utilize a novel architecture using a shared global attention block which obtains the benefits of hybrid SSM-attention architectures while minimizing the parameters dedicated to attention. Moreover, we describe the two-phase training regime Zamba underwent and release both phase 1 (pretrained) and final (annealed) models. Zamba was trained on a relatively small budget of approximately $\$200$k and a team of $7$ researchers over the course of a month and approaches leading models in performance, thus demonstrating that approaching the state-of-the-art in LLM pretraining does not necessarily require vast budgets or teams and is not restricted only to a few leading companies.

It is an interesting question what the source of the remaining gap between Zamba and the leading models at this scale is. The Zamba model trained until the end of phase 1 achieves Llama2 levels of performance from only 1T tokens, while Llama2 was trained on at least 2T tokens. This difference could arise from dataset differences, although given that Zamba's dataset is simply comprised of deduplicated open web datasets, it would be unlikely for Zamba's dataset to be significantly superior in quality to Llama2. It is also possible that our architecture gives us significant advantages over the Llama transformer architecture and indeed, despite claims that Mamba models struggle with ICL \citep{park2024mamba}, even our base model performs comparably to Llama2 here on MMLU, an evaluation metric which is known to require significant ICL to perform well. It is thus possible that even a single attention layer is enough to reach transformer parity on ICL. 

There is then the question of how to close the gap between Zamba and the leading open-weight models such as Mistral, Gemma, and Llama3. It is likely that some fraction of this gap is caused by a large disparity in the number of pretraining tokens. Our model is trained on only 1T tokens vs 8T for Gemma, 15T for Llama3, and an unknown (but likely similarly large) quantity for Mistral. Like others, and in accordance with the Chinchilla scaling laws, we observe continuing increases in performance (on a log scale) even towards the end of training, implying that our model has not plateaued in loss and could usefully be trained on many more tokens. Pretraining dataset quality is likely also a significant factor given that for our dataset, while utilizing existing well-known web datasets, we performed only straightforward filtering and deduplication, while there are many techniques for significantly improved dataset preparation available in the literature \citep{xie2024doremi,tirumala2024d4,maini2024rephrasing}. 

We see significant improvements on many evaluation scores during the annealing phase, providing an independent replication of the claims of miniCPM \citep{hu2024minicpm}. Given the significant jumps between base pretrained models such as Llama1, Llama2, and OLMo with relatively well-known datasets, and leading models in the 7B range such as Mistral and Gemma, it is likely that the augmentation of web data with synthetic, instruct, or other high quality data-sources played a key part in the performance of these models. This could either be during pretraining itself, as argued for by the phi series of models \citep{li2023textbooks,abdin2024phi}, or in an annealing phase similar to miniCPM and the one we performed. Empirically, we see that Zamba's annealing phase closes about half or more of the gap from Llama2 to state-of-the-art models especially on more reasoning-like evaluations such as MMLU~\citep{mmlu} and ARC~\citep{arc}. It is possible that an improved annealing phase on significantly more synthetic data, or pretraining on significantly more tokens may close this gap. 

While several recent works have claimed to perform annealing, relatively little has yet been published on the exact methods used for this or the performance of the base model before annealing began. We demonstrate that annealing on high quality data can significantly improve a model from approximately a Llama2 level base to closing the gap with leading models. Understanding the extent to which this performance improvement is real and enhances the model's true capabilities or whether it is simply training the model to respond in a more evaluation-friendly manner remains to be elucidated. Moreover, many questions remain open such as the optimal annealing schedule, the optimal replay fraction of original pretraining data, and the optimal composition of the annealing datasets. We hope that by releasing our base model, and all annealing checkpoints, we can help the academic community begin answering these questions. 

With Zamba, by both validating the performance of Mamba at scale and by pioneering a novel architecture beyond that, we have also taken a step towards moving away from the standard transformer architecture for training state-of-the-art models. We believe that architectural innovations are relatively understudied given the success of the scaling paradigm. However, as the cost of training state-of-the-art models increases, and as the benefits of scaling data at the small scale become increasingly marginal, the possibility of constant gains from improved architectures and pretraining paradigms vs logarithmic ones from additional data become increasingly important in pushing forward the frontier in performance for smaller model size brackets.

Finally, with Zamba, we make available all checkpoints during both the pretraining and annealing phase (one every 2500 steps) for scientific study. We believe that many scientific questions of deep interest for understanding the learning dynamics of such models can only be approached by studying the evolution of model weights during training and few academic groups have the budget or expertise to train models close to the state of the art. As such, we encourage other groups who train models at the frontier to consider making available and open the checkpoints of their models as well.

\clearpage

\section*{Author Contributions}
\textbf{Paolo} — Contributed to core infrastructure. Lead annealing experiments. Lead evaluations. Lead HuggingFace conversion and release. Contributed to architecture search experiments.

\textbf{Quentin} — Contributed to core infrastructure. Lead infrastructure development and training optimisation. Lead inference optimization. Contributed to cluster management and maintenance. Contributed to evaluations.

\textbf{Yury} — Contributed to core infrastructure. Lead dataset preparation and processing. Contributed to cluster management and maintenance. Contributed to evaluations.

\textbf{James} — Contributed to architecture search experiments. Invented final architecture used in the paper.

\textbf{Jonathan} — Contributed to architecture search experiments. %First proposed and experimented with an SSM - Transformer hybrid architecture.

\textbf{Adam} — Advised overall the project. Co-lead the curriculum approach. Contributed to annealing experiments.

\textbf{Beren} — Overall project lead. Contributed to core infrastructure. Contributed to annealing experiments. Lead creation and processing of annealing datasets. Contributed to architecture search experiments. Contributed to evaluations. Primary author of the technical report.

\section*{Acknowledgements}

We would like to acknowledge the rest of the Zyphra team for their support, including Steven Brook, Nick Alonso, Vasudev Shyam, Anna Golubeva, Tomas Figliolia and Krithik Puthalath for helpful discussions and feedback.

\bibliographystyle{apalike}
\bibliography{main}

\clearpage
%\section*{Appendix}
\appendix

\section*{Model and Training Hyperparameters}
\label{model_training_hyperparams_section}

\begin{table}[H]
\begin{tabular}{
>{\columncolor[HTML]{EFEFEF}}c c}
\cellcolor[HTML]{C0C0C0}\textbf{Hyperparameter} & \cellcolor[HTML]{C0C0C0}\textbf{Value} \\
Number of Layers                                & 80                                     \\
Hidden Dimension                                & 3712                                   \\
State Dimension                                 & 16                                     \\
Convolution Dimension                           & 4                                      \\
Number of Attention Heads                       & 16                                     \\
Context Length                                  & 4096                                   \\
Batch Size                                      & 512                                    \\
Max Learning Rage                               & 1.5e-4                                 \\
LR Decay Schedule                               & Cosine                                 \\
Minimum LR                                      & 7.5e-5                                 \\
Weight Decay                                    & 0.1                                    \\
Adam Beta2                                      & 0.95                                   \\
LR Warmup                                       & 0.01                                   \\
Gradient Clipping                               & 1.0                                    \\
Training Precision                              & BF16                                  
\end{tabular}
\end{table}

%\section*{Evaluations through time}

%\section*{Annealing Results}

%// add example loss and eval curves through time -- espcially discussion of the double descent of the validation loss in case anybody finds it interesting

%\section*{Ablations}

\end{document}